\pdfoutput=1
\documentclass[11pt]{article}

\usepackage[final]{nlp4MusA}

\usepackage{times}
\usepackage{latexsym}
\usepackage{listings}
\usepackage{subcaption}
\usepackage{caption}

\usepackage[T1]{fontenc}
\usepackage[utf8]{inputenc}

\usepackage{microtype}

\usepackage{inconsolata}

\usepackage{graphicx}

\title{Musical ethnocentrism in Large Language Models}

\author{Anna Kruspe \\
  Munich University of Applied Sciences \\
  Lothstr. 64, 80335 Munich, Germany \\ 
  \texttt{anna.kruspe@hm.edu} }

\begin{document}
\maketitle
\begin{abstract}
Large Language Models (LLMs) reflect the biases in their training data and, by extension, those of the people who created this training data. Detecting, analyzing, and mitigating such biases is becoming a focus of research. One type of bias that has been understudied so far are geocultural biases. Those can be caused by an imbalance in the representation of different geographic regions and cultures in the training data, but also by value judgments contained therein.

In this paper, we make a first step towards analyzing musical biases in LLMs, particularly ChatGPT and Mixtral. We conduct two experiments. In the first, we prompt LLMs to provide lists of the ``Top 100'' musical contributors of various categories and analyze their countries of origin. In the second experiment, we ask the LLMs to numerically rate various aspects of the musical cultures of different countries. Our results indicate a strong preference of the LLMs for Western music cultures in both experiments.
\end{abstract}

\section{Introduction}
It has long been known that machine learning models pick up and thus perpetuate human biases in various ways, most prominently by learning them from their training data. For text-based models, even early embedding approaches exhibited e.g. gender bias \cite{bolukbasi2016man}. With the recent rise of Large Language Models (LLMs), gender and race biases were quickly discovered and analyzed in various domains \cite{kotek2023gender,sun2023aligning,omiye2023large,warr2023implicit}. Types of bias that have been considered somewhat less include those based on culture and geography. However, \cite{manvi2024large} recently showed that LLMs also exhibit those, both implicitly and explicitly. Their research demonstrated that when prompted to rate random locations on Earth on various characteristics, LLMs generally yielded lower ratings for certain regions, e.g. the global South. There appear to be correlations with the coverage of regions and their cultural and historical significance in the training data, statistics across a range of aspects around the world, and possibly structural biases of the institutions creating these models, which are mainly based in North America and Europe.

We hypothesize that such biases are not only present for purely geographic topics, but also for cultural developments in different regions of the world. In this paper, we conduct first experiments to detect such biases with regards to music culture. Those are based on two different types of measurements, translated into prompts: a) Asking models to give an overview of top musical artists, and b) asking models to rate aspects of musical culture in different regions of the world. The first experiment elicits model bias on an open-ended question with regards to presence of different cultural regions in the models, while the second one employs a direct comparison to extract implicit judgments learned by the models. To gain insights into the influence of where and how the model was trained, we prompt two models from different regions of the world in four languages.

The rest of the paper is structured as follows: Section \ref{sec:related} gives an overview of other work in the field of geocultural biases in LLMs. Section \ref{sec:methodology} provides details of our experimental design. The results are presented in section \ref{sec:res}. Finally, sections \ref{sec:discussion} and \ref{sec:future} discuss our findings and make suggestions for future work.

\section{Related work}\label{sec:related}
Initial studies, such as those discussed in \cite{2023culturalbias} and \cite{2023beerprayer}, show LLMs often encode biases favoring Western, English-speaking norms, impacting their fairness and representation of non-Western cultures as well as performance on non-Western topics, like Traditional Chinese Medicine \cite{2024tcm}.

%Recent advances also highlight geographic biases in LLMs, demonstrating the importance of spatial and geographic knowledge encoded in these models. This knowledge, while enhancing geospatial applications and GeoAI frameworks, is often fraught with biases that skew data representation, favoring urban and affluent areas over rural or economically disadvantaged ones. Studies testing LLMs in tasks like route planning and geocoded information retrieval reveal these models' limitations in abstract reasoning and raise concerns about their reliance on memorization, which often leads to significant regional disparities in knowledge and capabilities \cite{}.

Further research seeks methodologies to measure and mitigate these biases more accurately. The interdisciplinary approach in \cite{2024beyondhuman} and the survey on modeling culture in LLMs \cite{2024measuringculture} propose new frameworks for understanding and adjusting the embedded cultural values in LLMs. The ``CulturePark'' \cite{culturepark2024} initiative and the ``NormAd'' \cite{normad2024} benchmark are notable in their attempts to simulate cross-cultural communication scenarios and assess LLMs' adaptability to cultural contexts through synthetic story generation, providing novel approaches to evaluating cultural sensitivity and adaptability in AI technologies. The ``CDEval'' \cite{cdeval2023} benchmark specifically addresses the need to evaluate the cultural dimensions of LLMs, integrating automated generation and human verification to assess cultural traits across multiple domains. Similarly, the ``CultureLLM'' \cite{2023culturellm} project aims to fine-tune LLMs on culturally diverse data.

In the music domain, \cite{2024musicmaestro} illustrates domain-specific biases, arguing for more comprehensive benchmarks in varied knowledge domains.

\section{Methodology}\label{sec:methodology}
In this section, we will describe our experimental design, including used models, prompt design, prompted tasks, and postprocessing of the results.

\subsection{Models}
We tested our bias prompts on two different models via their online interfaces:
\begin{itemize}
\item ChatGPT-4 (paid version) via its online interface (\url{https://chatgpt.com/})
\item Mixtral-8x7B via the online interface under \url{https://deepinfra.com/mistralai/Mixtral-8x7B-Instruct-v0.1}, maximum new token length set to 10,000
\end{itemize}
ChatGPT was created in the US by OpenAI, while Mixtral was released by the French company Mistral AI. We wanted to compare models from two different regions of the world on this geocentric task. A more geographically wide-ranging selection of LLMs would be of high interest for future comparisons. Currently, Chinese institutions are also intensifying their efforts in the LLM domain, but we were not able to obtain access to a freely available Chinese model.

\begin{lstlisting}[language={}, numbers=none, xleftmargin=0pt,breaklines=true, label=lis:prompt_en, caption=Example prompt for the rating experiment] 
You will be given a country randomly sampled from all human-populated locations on Earth. You give your rating keeping in mind that it is relative to all other human-populated locations on Earth (from all continents, countries, etc.). You provide ONLY your answer in the exact format "My answer is X.X." where `X.X' represents your rating for the given topic.

...

task: Agreeableness of music
region: Denmark
\end{lstlisting}

\subsection{Prompt design}
We conducted two experiments. In the first one, we asked open-ended questions about the ``Top 100'' musical performers of various types. Those included bands, solo musicians, singers, instrumentalists, and composers. Prompts were simply of the form \lstinline{"Name the Top 100 singers/instrumentalists/bands/..."}. We then asked the model to extend this list with the performers' countries of origin.

In the second experiment, we asked the models to rate certain characteristics of the music of all countries in the world. We used the methodology from \cite{manvi2024large} to design our prompts. In essence, the LLMs are given an over-all task of providing ratings on a certain topic for a certain region of the world compared to all other inhabited areas. An example is shown in Listing \ref{lis:prompt_en}. Then, we focused on a specific characteristic and gave the model a list of all countries. For such subjective topics, it is usually not possible to ask for direct comparisons (e.g. ``which country has the best music'') due to content filters, but ratings worked well. The aspects of music culture included agreeableness, successfulness, musical creativity, global influence, musical tradition, and musical complexity.

Prompts were designed in English, and then translated to Spanish, Chinese, and French using ChatGPT-4. The list of countries was kept in English. ChatGPT was prompted in English, Spanish, and Chinese, and Mixtral was prompted in English and French. Each experiment was repeated three times on each model and each language to account for different initializations. 

%\newpage

\subsection{Postprocessing}
For the Top 100 experiment, we then calculated the frequency of each country's appearance across all runs. In cases where the model named multiple countries of origin for one performer, we only kept the first one for simplicity.

The rating results were normalized by mean and standard deviation for each characteristic. Then, we averaged those normalized results across all three runs for each characteristic.

Our full results and analysis notebooks are available on \url{https://github.com/annakaa/musical_ethnocentrism}.

\section{Results} \label{sec:res}
\subsection{``Top 100'' results}\label{sec:res_top}
\begin{figure*}
    \centering
    \includegraphics[width=1\linewidth, trim=0 50 0 0, clip]{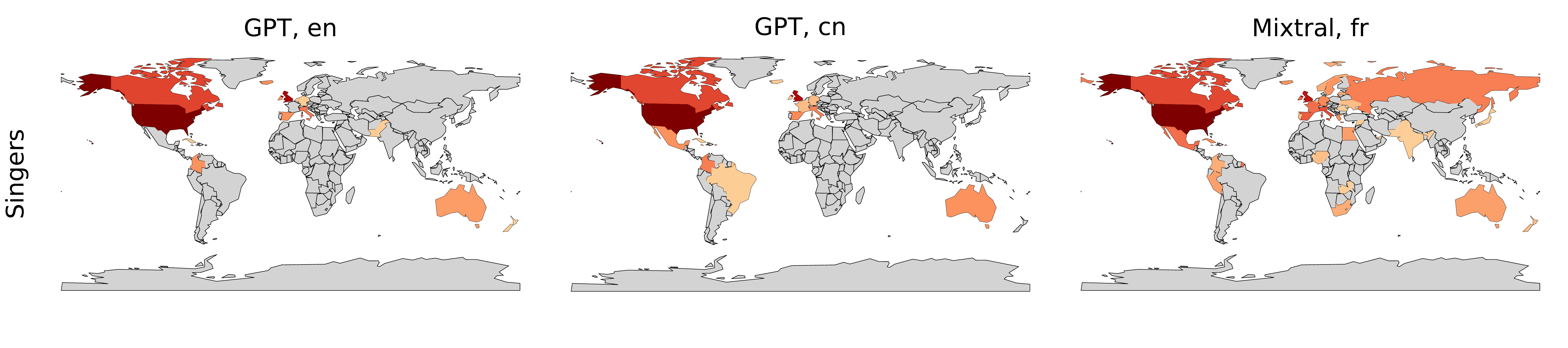}
    \includegraphics[width=0.3\linewidth]{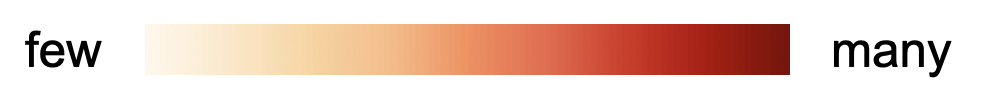}
    %\captionsetup{skip=0.1pt}
    \caption{Example results of the ``Top 100'' experiments for singers, prompted on GPT in English and Chinese, and on Mixtral in French. Gray means None, and darker colors indicate higher numbers.}
    
    \label{fig:res_top100_ex}
\end{figure*}
%add: Bands Mixtral, fr
An example result for the ``Top 100'' experiments is shown in Figure \ref{fig:res_top100_ex}, with the full results in Figure \ref{fig:res_top100} (appendix).
As hypothesized, the results are very focused on Western countries, especially the U.S. South American representation varies a bit, whereas Asia and Africa are completely underrepresented. The effect is particularly strong for bands and singers. For the question about solo artists and instrumentalists, results are a bit more diverse. The prompt about composers has a stronger European focus, but also results in a surprisingly high number of those from the U.S. (including some who are possibly lesser-known in the rest of the world).

When prompting with different models and different languages, the results vary, but are somewhat inconclusive. Spanish-language prompts do seem to lead to a slightly stronger representation of Spain and South America, and Chinese-language ones to a stronger focus on China, but none of the changes are very pronounced. Compared to ChatGPT, Mixtral appears to produce slightly more diverse results, especially with regards to Africa (interestingly, though, more for the countries with English as their official language rather than French).

\subsection{Rating results}\label{sec:res_ratings}
\begin{figure*}
    \centering
    \includegraphics[width=1\linewidth,  trim=0 50 0 0, clip]{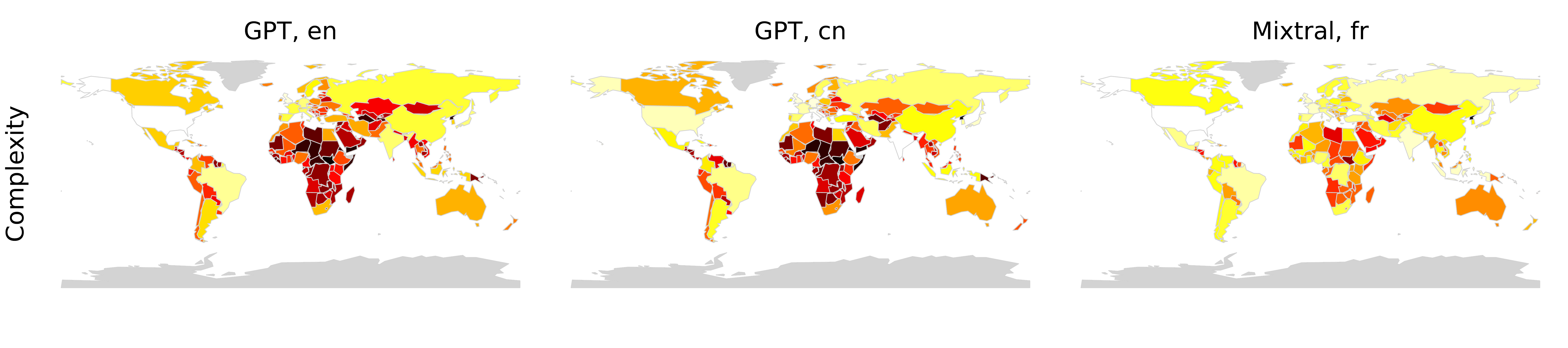}
    \includegraphics[width=0.3\linewidth]{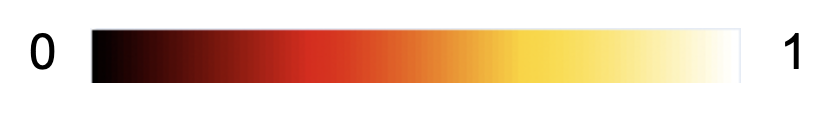}
        %\captionsetup{skip=0.1pt}

    \caption{Example results of the rating experiments for musical complexity, prompted on GPT in English and Chinese, and on Mixtral in French. Scale runs from dark red (low rating) to bright yellow (high rating).}
    \label{fig:res_rating_ex}
\end{figure*}
% Complexity GPT, en and Mixtral, Fr
An example result of the experiments where we asked LLMs to rate aspects of music culture in different countries is presented in Figure \ref{fig:res_rating_ex}, and the full results are shown in Figure \ref{fig:res_rating} (appendix).
Once again, we see a strong tendency towards Western countries, especially the U.S. Correlating with the results of the previous experiments, Asian and African countries are rated much lower in comparison, while South America lies somewhere in the middle. This is true for almost all prompted aspects. The outlier appears to be ``Tradition'', where, for example, India tends to be rated higher. This may happen due to training data sources that are more focused on folkloristic (``world'') music rather than pop or classical music, which may become associated with the ``Tradition'' keyword.

Once again, we do no see major effects between models and languages. Prompting in Chinese appears to emphasize the U.S. and India, but not necessarily China itself, whereas prompting in Spanish once again leads to slightly higher ratings for Spain and South America. When using Mixtral, we once again obtain somewhat more balanced results. In particular, the ``Tradition'' prompt yields higher ratings in Africa, and this time mainly for French-speaking countries. This may happen due to a higher frequency of French-language sources in Mixtral training.

\section{Discussion}\label{sec:discussion}
As expected, we observed a strong dominance of the Western world, particularly the U.S., in both tasks. South America was comparatively well-represented, whereas Asia and Africa were almost never mentioned in the ``Top 100'' experiments, and rated consistently lower in the second experiment.

Both models produce slightly different results, with Mixtral appearing a bit more diverse. However, there is some indication that state-of-the-art language models are trained on most of the text data currently available on the internet, meaning that the cultural distribution of training data may not vary too much between any current models\footnote{\url{https://situational-awareness.ai/from-gpt-4-to-agi/\#The_data_wall}}.

The language in which prompts are given to the model does appear to play a role, but not in a very straightforward way (e.g. Chinese-language prompting does not lead to China being mentioned significantly more often). Due to the cross-lingual abilities of LLMs, the language of the context may in fact play a smaller role than language in the training data. CommonCrawl, often named as the biggest source of LLM training data, contains around 46\% English-language text, which the second-most frequent language being Russian at just 6\%\footnote{\url{https://commoncrawl.github.io/cc-crawl-statistics/plots/languages}}. Nevertheless, the language of a country will in all probability be implicitly somewhat more strongly associated with its culture.

These results may not seem like a big issue at first glance, but could lead to undesired effects in downstream tasks, e.g. when used in recommendation pipelines or for assistance in writing about culture. This is particularly insidious because a) these biases are then much harder to detect, and b) they will lead to an amplification of biases already present in existing, human-produced material. On the other hand, users may in fact expect models to behave the way they currently do, especially when considering the ``Top 100'' experiment. The question then becomes whether future models should maintain those biases, or could potentially serve to offer a more diverse and educative view of the world to their users. This will become a more pressing consideration with future individualization of models.

\section{Future work}\label{sec:future}
In this work, we presented a first step towards detecting cultural biases in LLMs with a focus on the music domain. As mentioned above, it would be very interesting to see whether LLMs from other parts of the world (first and foremost China) perpetuate the same biases. Future work could also analyze other music-related tasks around the world, and compare with other aspects of culture. On a smaller note, the results were obtained via the online interfaces of the models which may filter or change results; future work could employ the models directly for more control.

Beyond analyzing these biases, an important research goal lies in mitigating them. When considering the ``Top 100'' task, this is very subjective. An interesting research direction may be aimed more towards human-computer interaction: What do users expect when prompting models for recommendations like these? From an ethical standpoint, should models then fulfil users' expectations, or aim for more diversity than what a human author or the training data may provide? Answers may lie in integrating external knowledge sources (e.g. knowledge graphs) into LLMs, but also in adapting them towards individual users.

For the ratings task, possible solutions are much harder to determine. In principle, the whole task of rating musical cultures is not well-posed, but it reveals underlying judgments learned by the model, which may influence downstream tasks (including the ``Top 100'' experiment). Removing these judgments may be impossible as they appear to be implicit in the training data. A possible future direction may lie in making these influences more transparent to users, allowing them to decide for themselves whether the model's answer is based on the correct assumptions \cite{kruspe2024towards}.

\newpage
% Bibliography entries defined in nlp4MusA.bib
\bibliography{nlp4MusA}
\newpage
\appendix

\section{Appendix}
\label{sec:appendix}
\smallskip
\begin{minipage}{\textwidth}
    \centering
    \includegraphics[width=1.4\hsize, angle=270]{fig/TOP100.pdf}
    \vspace{0cm} 
    \includegraphics[width=0.3\linewidth]{fig/orrd_label.png}
    
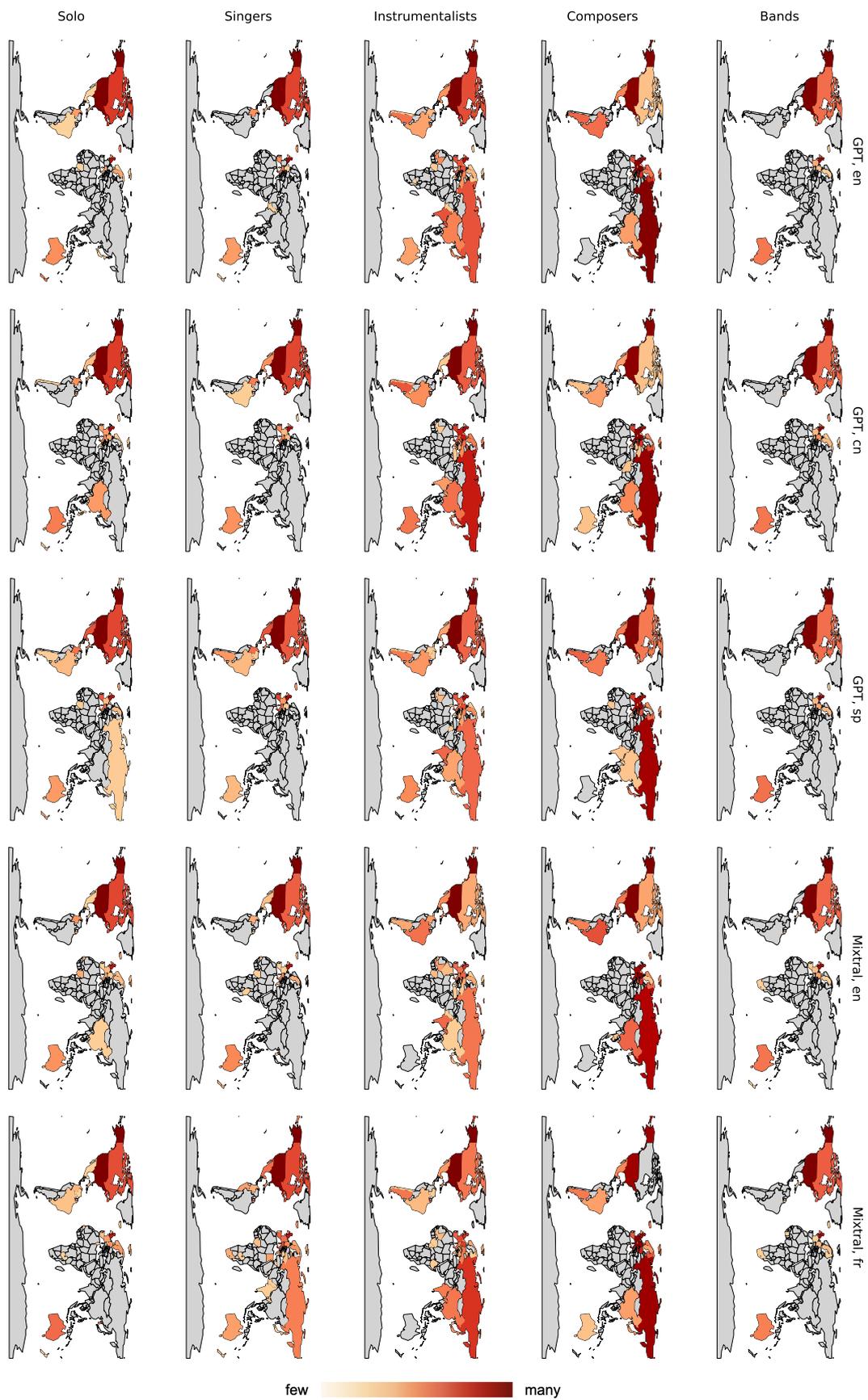
\captionof{figure}{``Top 100'' result graphs. Gray means None, and darker colors indicate higher numbers.}
    \label{fig:res_top100}
\end{minipage}

\begin{figure*}
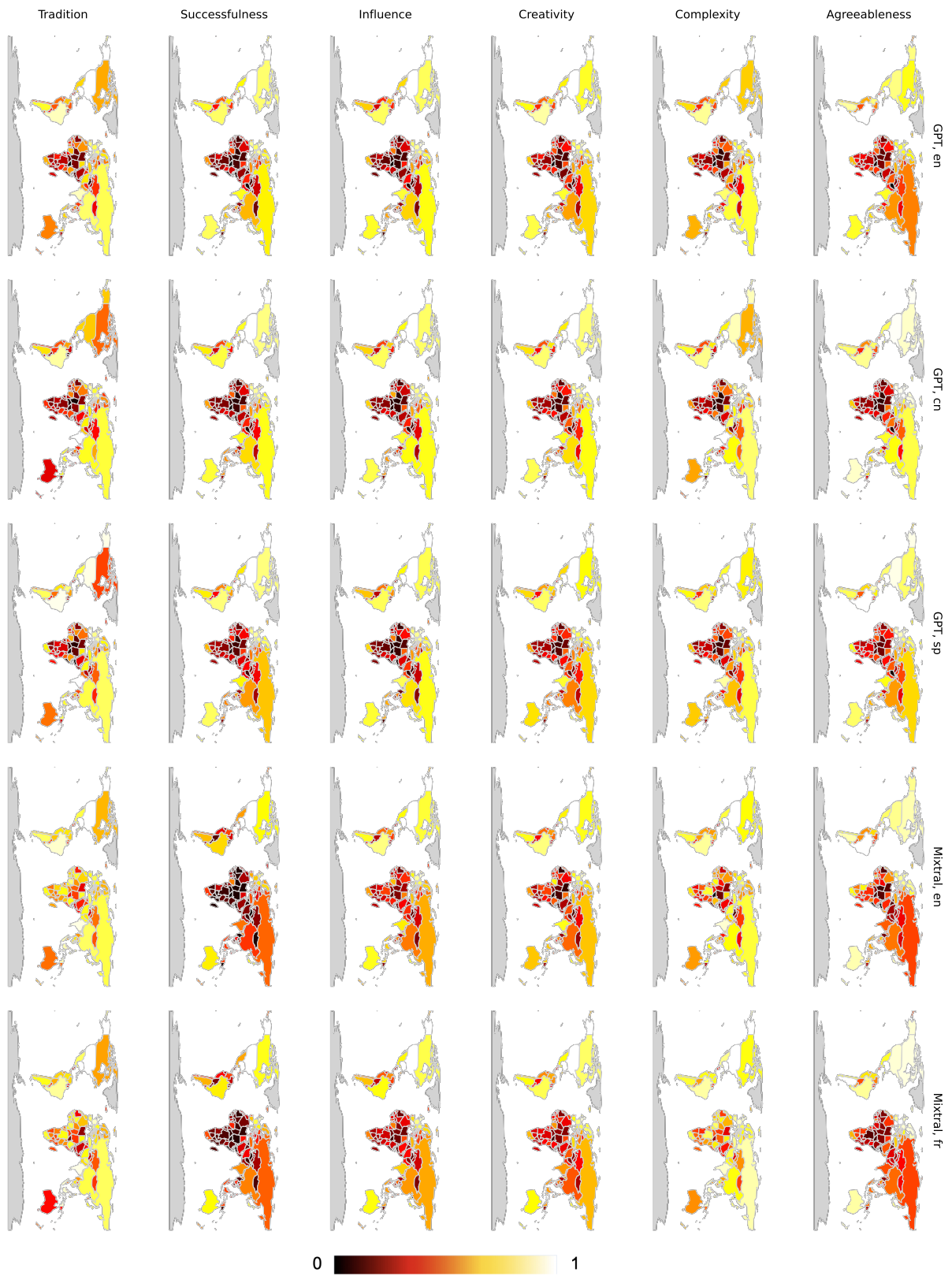

    \centering
    \includegraphics[width=1.3\textwidth, angle=270]{fig/PROPERTIES.pdf}
    \vspace{0cm} 
    \includegraphics[width=0.3\linewidth]{fig/hot_label.png}
    
    \caption{Rating result graphs. Scale runs from dark red (low rating) to bright yellow (high rating).}
    \label{fig:res_rating}
\end{figure*}
% \begin{figure}
%     \centering
%     \includegraphics[width=0.5\linewidth, angle=90]{fig/PROPERTIES2.png}
%     \caption{Enter Caption}
%     \label{fig:enter-label}
% \end{figure}
\end{document}